\def\year{2015}
\documentclass[letterpaper]{article}
\usepackage{aaai}
\usepackage{times}
\usepackage{helvet}
\usepackage{courier}
\usepackage{subcaption}
\usepackage{amsfonts}
\usepackage[percent]{overpic}
\usepackage{dsfont}
\begin{document}
\title{Deep Recurrent Q-Learning for Partially Observable MDPs}
\author{Matthew Hausknecht and Peter Stone\\
Department of Computer Science\\
The University of Texas at Austin\\
\{mhauskn, pstone\}@cs.utexas.edu\\
}
\maketitle

\begin{abstract}
\begin{quote}
Deep Reinforcement Learning has yielded proficient controllers for
complex tasks. However, these controllers have limited memory and rely
on being able to perceive the complete game screen at each decision
point. To address these shortcomings, this article investigates the
effects of adding recurrency to a Deep Q-Network (DQN) by replacing
the first post-convolutional fully-connected layer with a recurrent
LSTM. The resulting \textit{Deep Recurrent Q-Network} (DRQN), although
capable of seeing only a single frame at each timestep, successfully
integrates information through time and replicates DQN's performance
on standard Atari games and partially observed equivalents featuring
flickering game screens. Additionally, when trained with partial
observations and evaluated with incrementally more complete
observations, DRQN's performance scales as a function of
observability. Conversely, when trained with full observations and
evaluated with partial observations, DRQN's performance degrades less
than DQN's. Thus, given the same length of history, recurrency is a
viable alternative to stacking a history of frames in the DQN's input
layer and while recurrency confers no systematic advantage when learning
to play the game, the recurrent net can better adapt at evaluation
time if the quality of observations changes.
\end{quote}
\end{abstract}

\section{Introduction}
Deep Q-Networks (DQNs) have been shown to be capable of learning
human-level control policies on a variety of different Atari 2600
games \cite{mnih15}. True to their name, DQNs learn to estimate the
Q-Values (or long-term discounted returns) of selecting each possible
action from the current game state. Given that the network's Q-Value
estimate is sufficiently accurate, a game may be played by selecting
the action with the maximal Q-Value at each timestep. Learning
policies mapping from raw screen pixels to actions, these networks
have been shown to achieve state-of-the-art performance on many Atari
2600 games.

However, Deep Q-Networks are limited in the sense that they learn a
mapping from a limited number of past states, or game screens in the
case of Atari 2600. In practice, DQN is trained using an input
consisting of the last four states the agent has encountered. Thus DQN
will be unable to master games that require the player to remember
events more distant than four screens in the past. Put differently,
any game that requires a memory of more than four frames will appear
non-Markovian because the future game states (and rewards)
depend on more than just DQN's current input. Instead of a Markov
Decision Process (MDP), the game becomes a Partially-Observable Markov
Decision Process (POMDP).

\begin{figure}[t!]
\begin{subfigure}{.15\textwidth}
  \centering
  \centering \includegraphics[width=\textwidth]{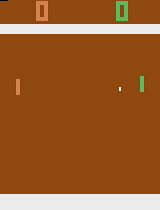}
  \caption{Pong}
  \label{fig:pong}
\end{subfigure}
\begin{subfigure}{.15\textwidth}
  \includegraphics[width=\textwidth]{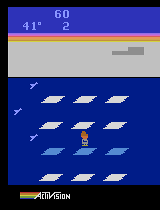}
  \caption{Frostbite}
  \label{fig:frostbite}
\end{subfigure}
\begin{subfigure}{.15\textwidth}
  \includegraphics[width=\textwidth]{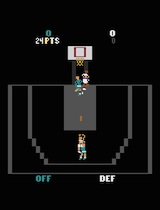}
  \caption{Double Dunk}
  \label{fig:double_dunk}
\end{subfigure}
\caption{Nearly all Atari 2600 games feature moving objects. Given only one frame of input, Pong, Frostbite, and Double Dunk are all POMDPs because a single observation does not reveal the velocity of the ball (Pong, Double Dunk) or the velocity of the icebergs (Frostbite).}
\label{fig:atari_games}
\end{figure}

Real-world tasks often feature incomplete and noisy state information
resulting from partial observability. As Figure \ref{fig:atari_games}
shows, given only a single game screen, many Atari 2600 games are
POMDPs. One example is the game of Pong in which the current screen
only reveals the location of the paddles and the ball, but not the
velocity of the ball. Knowing the direction of travel of the ball is a
crucial component for determining the best paddle location.

We observe that DQN's performance declines when given incomplete state
observations and hypothesize that DQN may be modified to better deal
with POMDPs by leveraging advances in Recurrent Neural
Networks. Therefore we introduce the \textit{Deep Recurrent Q-Network}
(DRQN), a combination of a Long Short Term Memory
(LSTM) \cite{hochreiter97} and a Deep Q-Network. Crucially, we
demonstrate that DRQN is capable of handing partial observability, and
that when trained with full observations and evaluated with partial
observations, DRQN better handles the loss of information than does
DQN. Thus, recurrency confers benefits as the quality of observations
degrades.

\subsection{Deep Q-Learning}
Reinforcement Learning \cite{sutton98} is concerned with learning
control policies for agents interacting with unknown
environments. Such environments are often formalized as a Markov
Decision Processes (MDPs), described by a 4-tuple
$(\mathcal{S},\mathcal{A},\mathcal{P},\mathcal{R})$. At each timestep
$t$ an agent interacting with the MDP observes a state
$s_t \in \mathcal{S}$, and chooses an action $a_t \in \mathcal{A}$
which determines the reward $r_t \sim \mathcal{R}(s_t,a_t)$ and next
state $s_{t+1} \sim \mathcal{P}(s_t,a_t)$.

Q-Learning \cite{watkins92} is a model-free off-policy algorithm for
estimating the long-term expected return of executing an action from a
given state. These estimated returns are known as Q-values. A higher
Q-value indicates an action $a$ is judged to yield better long-term
results in a state $s$. Q-values are learned iteratively by updating
the current Q-value estimate towards the observed reward plus the max
Q-value over all actions $a'$ in the resulting state $s'$:

\begin{equation}
Q(s,a) := Q(s,a) + \alpha \big( r + \gamma \max_{a'} Q(s',a') - Q(s,a) \big)
\end{equation}

Many challenging domains such as Atari games feature far too many
unique states to maintain a separate estimate for each
$\mathcal{S} \times \mathcal{A}$. Instead a model is used to
approximate the Q-values \cite{mnih15}. In the case of Deep
Q-Learning, the model is a neural network parameterized by weights and
biases collectively denoted as $\theta$. Q-values are estimated online
by querying the output nodes of the network after performing a forward
pass given a state input. Such Q-values are denoted
$Q(s,a|\theta)$. Instead of updating individual Q-values, updates are
now made to the parameters of the network to minimize a differentiable
loss function:

\begin{equation}
L(s,a|\theta_i) = \big( r + \gamma \max_{a'} Q(s',a'|\theta_i) - Q(s,a|\theta_i) \big)^2
\end{equation}

\begin{equation}
\theta_{i+1} = \theta_i + \alpha \nabla_\theta L(\theta_i)
\end{equation}

Since $|\theta| \ll |\mathcal{S} \times \mathcal{A}|$, the neural
network model naturally generalizes beyond the states and actions it
has been trained on. However, because the same network is generating
the next state target Q-values that are used in updating its current
Q-values, such updates can oscillate or
diverge \cite{tsitsiklis97}. Deep Q-Learning uses three techniques to
restore learning stability: First, experiences $e_t =
(s_t,a_t,r_t,s_{t+1})$ are recorded in a replay memory $\mathcal{D}$
and then sampled uniformly at training time. Second, a separate,
target network $\hat{Q}$ provides update targets to the main network,
decoupling the feedback resulting from the network generating its own
targets. $\hat{Q}$ is identical to the main network except its
parameters $\theta^-$ are updated to match $\theta$ every 10,000
iterations. Finally, an adaptive learning rate method such as
RMSProp \cite{tieleman12} or ADADELTA \cite{zeiler12} maintains a
per-parameter learning rate $\alpha$, and adjusts $\alpha$ according
to the history of gradient updates to that parameter. This step serves
to compensate for the lack of a fixed training dataset; the
ever-changing nature of $\mathcal{D}$ may require certain parameters
start changing again after having reached a seeming fixed point.

At each training iteration $i$, an experience $e_t =
(s_t,a_t,r_t,s_{t+1})$ is sampled uniformly from the replay memory
$\mathcal{D}$. The loss of the network is determined as follows:

\begin{equation}
L_i(\theta_i) = \mathbb{E}_{(s_t,a_t,r_t,s_{t+1}) \sim \mathcal{D}} \bigg[ \Big( y_i - Q(s_t,a_t;\theta_i)\Big)^2 \bigg]
\end{equation}

where $y_i = r_t + \gamma \max_{a'}\hat{Q}(s_{t+1},a';\theta^-)$ is the stale
update target given by the target network $\hat{Q}$. Updates performed
in this manner have been empirically shown to be tractable and
stable \cite{mnih15}.

\subsection{Partial Observability}
In real world environments it's rare that the full state of the system
can be provided to the agent or even determined. In other words, the
Markov property rarely holds in real world environments. A Partially
Observable Markov Decision Process (POMDP) better captures the
dynamics of many real-world environments by explicitly acknowledging
that the sensations received by the agent are only partial glimpses of
the underlying system state. Formally a POMDP can be described as a
6-tuple
$(\mathcal{S},\mathcal{A},\mathcal{P},\mathcal{R},\Omega,\mathcal{O})$. $\mathcal{S},\mathcal{A},\mathcal{P},\mathcal{R}$
are the states, actions, transitions, and rewards as before, except
now the agent is no longer privy to the true system state and instead
receives an observation $o \in \Omega$. This observation is generated
from the underlying system state according to the probability
distribution $o \sim \mathcal{O}(s)$. Vanilla Deep Q-Learning has no
explicit mechanisms for deciphering the underlying state of the POMDP
and is only effective if the observations are reflective of underlying
system states. In the general case, estimating a Q-value from an
observation can be arbitrarily bad since $Q(o,a|\theta) \neq
Q(s,a|\theta)$.

Our experiments show that adding recurrency to Deep Q-Learning allows
the Q-network network to better estimate the underlying system state,
narrowing the gap between $Q(o,a|\theta)$ and $Q(s,a|\theta)$. Stated
differently, recurrent deep Q-networks can better approximate actual
Q-values from sequences of observations, leading to better policies in
partially observed environments.

\section{DRQN Architecture}
\label{sec:arch}
\begin{figure}[ht!]
  \hspace*{-.5cm}
  \begin{overpic}[width=0.5\textwidth,tics=10]{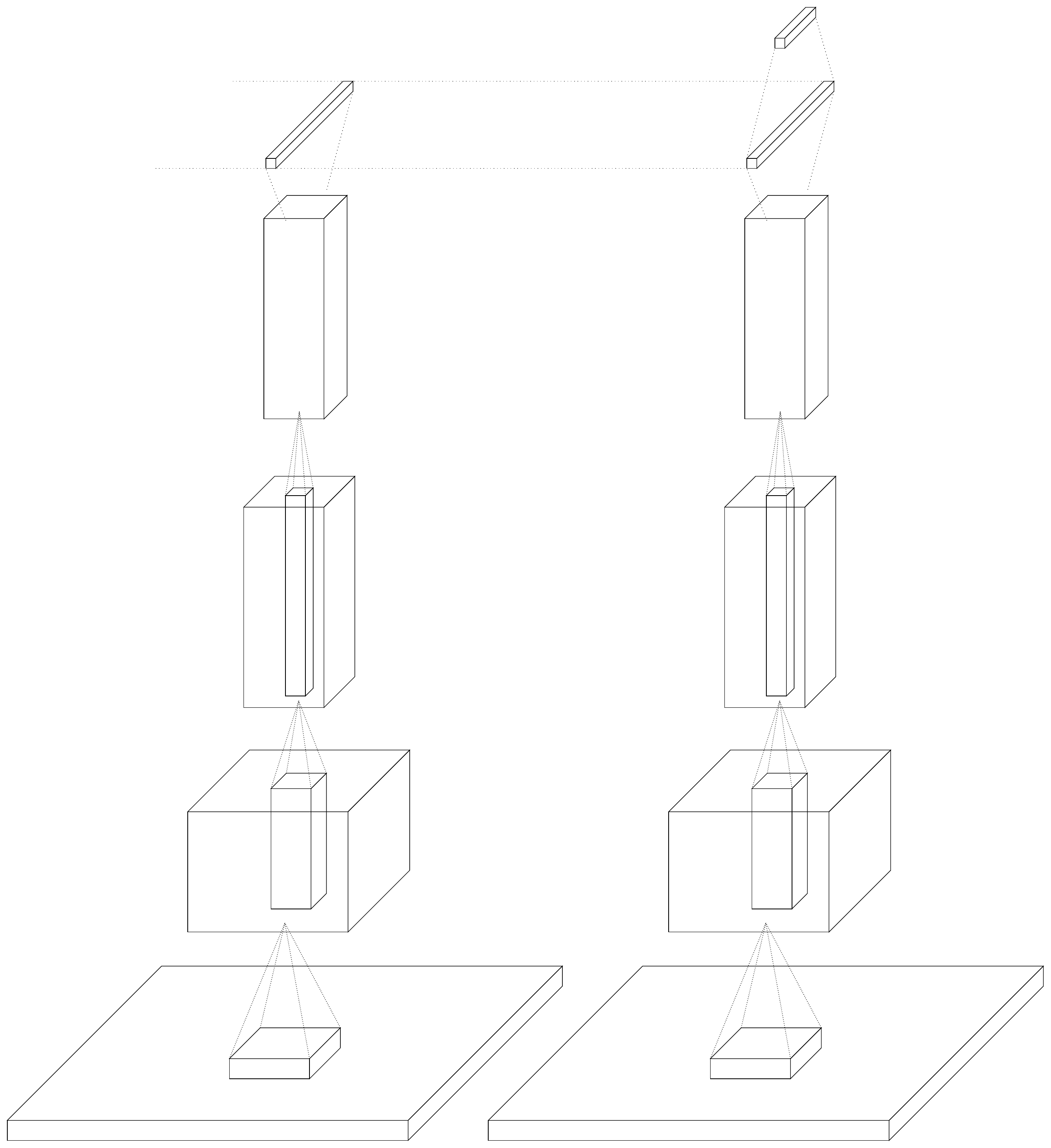}
    \put (43,97) {\textbf{Q-Values}}
    \put (73,97) {18}
    \put (43,88) {\textbf{LSTM}}
    \put (75,88) {512}
    \put (73,72) {\small{64}}
    \put (65,60) {\small{7}}
    \put (72,62) {\small{7}}
    \put (43,77) {\textbf{Conv3}}
    \put (43,73) {\small{64-filters}}
    \put (43,69) {\small{$3\times 3$}}
    \put (43,65) {\small{Stride 1}}
    \put (74,48) {\small{64}}
    \put (65,35) {\small{9}}
    \put (72,38) {\small{9}}
    \put (43,55) {\textbf{Conv2}}
    \put (43,51) {\small{64-filters}}
    \put (43,47) {\small{$4\times 4$}}
    \put (43,43) {\small{Stride 2}}
    \put (78,28) {\small{32}}
    \put (60,16) {\small{20}}
    \put (76,20) {\small{20}}
    \put (43,32) {\textbf{Conv1}}
    \put (43,28) {\small{32-filters}}
    \put (43,24) {\small{$8\times 8$}}
    \put (43,20) {\small{Stride 4}}
    \put (92,14) {\small{1}}
    \put (86,6) {\small{84}}
    \put (60,-2) {\small{84}}
    \put (18, 89) {\small{$\dots$}}
  \end{overpic}
  \caption{DRQN convolves three times over a single-channel image of the game screen. The resulting activations are processed through time by an LSTM layer. The last two timesteps are shown here. LSTM outputs become Q-Values after passing through a fully-connected layer. Convolutional filters are depicted by rectangular sub-boxes with pointed tops.}
  \label{fig:arch}
\end{figure}
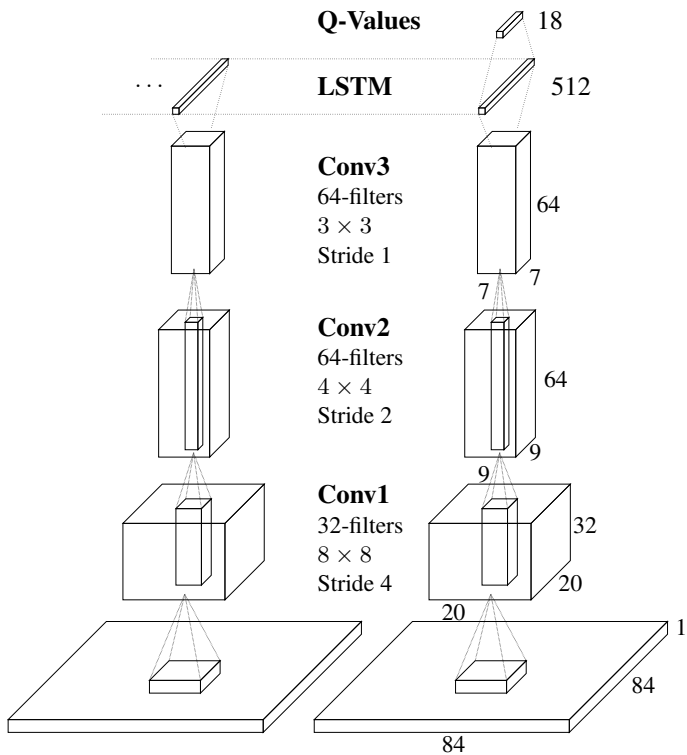

To isolate the effects of recurrency, we minimally modify the
architecture of DQN, replacing only its first fully connected layer
with a recurrent LSTM layer of the same size. Depicted in
Figure \ref{fig:arch}, the architecture of DRQN takes a single
$84 \times 84$ preprocessed image. This image is processed by three
convolutional layers \cite{lecun98} and the outputs are fed to the
fully connected LSTM layer \cite{hochreiter97}. Finally, a linear
layer outputs a Q-Value for each action. During training, the
parameters for both the convolutional and recurrent portions of the
network are learned jointly from scratch. We settled on this
architecture after experimenting with several variations; see Appendix
A for details.


\section{Stable Recurrent Updates}
Updating a recurrent, convolutional network requires each backward
pass to contain many time-steps of game screens and target
values. Additionally, the LSTM's initial hidden state may either be
zeroed or carried forward from its previous values. We consider two
types of updates:

\textbf{Bootstrapped Sequential Updates}: Episodes are selected
  randomly from the replay memory and updates begin at the beginning
  of the episode and proceed forward through time to the conclusion of
  the episode. The targets at each timestep are generated from the
  target Q-network, $\hat{Q}$. The RNN's hidden state is carried
  forward throughout the episode.

\textbf{Bootstrapped Random Updates}: Episodes are selected
  randomly from the replay memory and updates begin at random points
  in the episode and proceed for only \textit{unroll iterations}
  timesteps (e.g. one backward call). The targets at each timestep are
  generated from the target Q-network, $\hat{Q}$. The RNN's initial
  state is zeroed at the start of the update.

Sequential updates have the advantage of carrying the LSTM's hidden
state forward from the beginning of the episode. However, by sampling
experiences sequentially for a full episode, they violate DQN's random
sampling policy.

Random updates better adhere to the policy of randomly sampling
experience, but, as a consequence, the LSTM's hidden state must be
zeroed at the start of each update. Zeroing the hidden state makes it
harder for the LSTM to learn functions that span longer time scales
than the number of timesteps reached by back propagation through time.

Experiments indicate that both types of updates are viable and yield
convergent policies with similar performance across a set of
games. Therefore, to limit complexity, all results herein use the
randomized update strategy. We expect that all presented results would
generalize to the case of sequential updates.

Having addressed the architecture and updating of a Deep Recurrent
Q-Network, we now show how it performs on domains featuring partial
observability.

\section{Atari Games: MDP or POMDP?}
The state of an Atari 2600 game is fully described by the 128 bytes of
console RAM. Humans and agents, however, observe only the
console-generated game screens. For many games, a single game screen
is insufficient to determine the state of the system. DQN infers the
full state of an Atari game by expanding the state representation to
encompass the last four game screens. Many games that were previously
POMDPs now become MDPs. Of the 49 games investigated by \cite{mnih15},
the authors were unable to identify any that were partially observable
given the last four frames of input.\footnote{Some Atari games are
undoubtedly POMDPs such as Blackjack in which the dealer's cards are
hidden from view. Unfortunately, Blackjack is not supported by the ALE
emulator.} Since the explored games are fully observable given four
input frames, we need a way to introduce partial observability without
reducing the number of input frames given to DQN.

\section{Flickering Atari Games}
To address this problem, we introduce the \textit{Flickering Pong}
POMDP - a modification to the classic game of Pong such that at each
timestep, the screen is either fully revealed or fully obscured with
probability $p=0.5$. Obscuring frames in this manner probabilistically
induces an incomplete memory of observations needed for Pong to become
a POMDP.

In order to succeed at the game of Flickering Pong, it is necessary to
integrate information across frames to estimate relevant variables
such as the location and velocity of the ball and the location of the
paddle. Since half of the frames are obscured in expectation, a
successful player must be robust to the possibility of several
potentially contiguous obscured inputs.

\begin{figure}[htp]
\begin{subfigure}{.5\textwidth}
 \includegraphics[width=\textwidth]{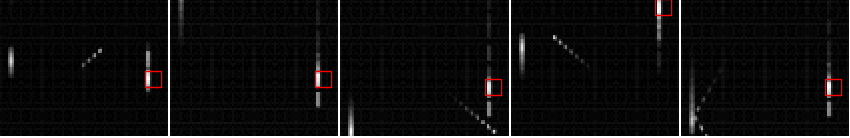}
\end{subfigure}
\begin{subfigure}{.5\textwidth}
 \includegraphics[width=\textwidth]{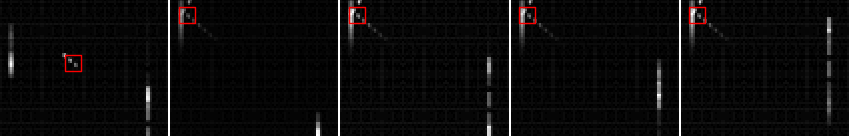}
 \caption{Conv1 Filters}
\end{subfigure}
\begin{subfigure}{.5\textwidth}
  \includegraphics[width=\textwidth]{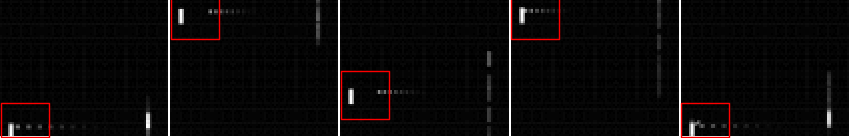}
\end{subfigure}
\begin{subfigure}{.5\textwidth}
  \includegraphics[width=\textwidth]{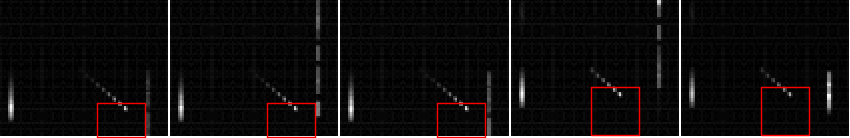}
 \caption{Conv2 Filters}
\end{subfigure}
\begin{subfigure}{.5\textwidth}
  \includegraphics[width=\textwidth]{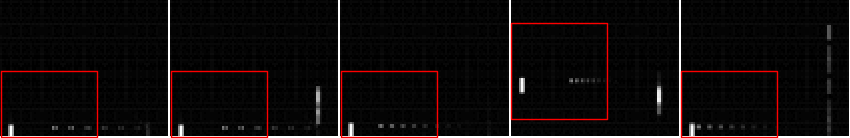}
\end{subfigure}
\begin{subfigure}{.5\textwidth}
  \includegraphics[width=\textwidth]{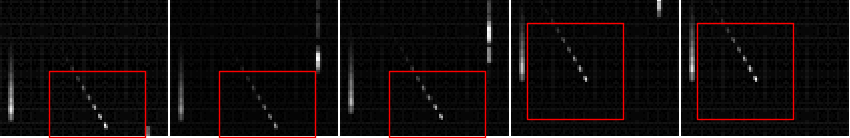}
\end{subfigure}
\begin{subfigure}{.5\textwidth}
  \includegraphics[width=\textwidth]{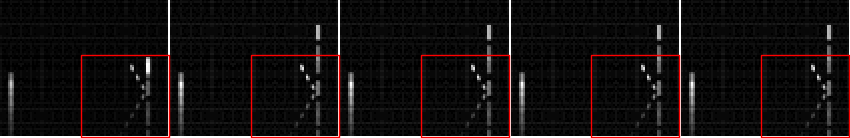}
 \caption{Conv3 Filters}
\end{subfigure}
\begin{subfigure}{.5\textwidth}
 \includegraphics[width=\textwidth]{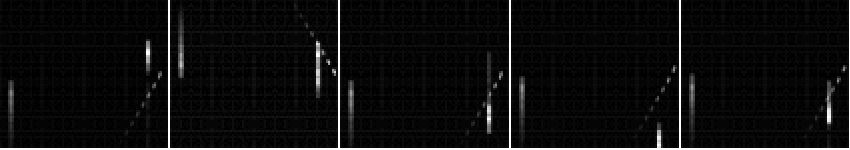}
\end{subfigure}
\begin{subfigure}{.5\textwidth}
 \includegraphics[width=\textwidth]{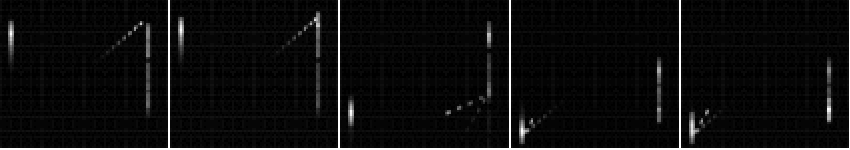}
\end{subfigure}
\begin{subfigure}{.5\textwidth}
 \includegraphics[width=\textwidth]{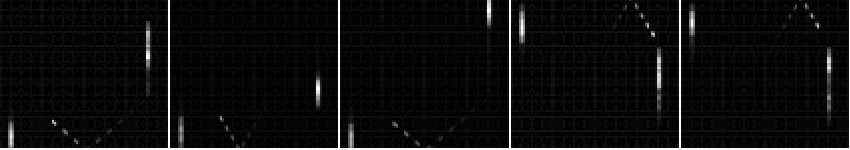}
 \caption{Image sequences maximizing three sample LSTM units}
 \label{fig:lstm_units}
\end{subfigure}
\caption{Sample convolution filters learned by 10-frame DQN on the game of Pong. Each row plots the input frames that trigger maximal activation of a particular convolutional filter in the specified layer. The red bounding box illustrates the portion of the input image that caused the maximal activation. Most filters in the first convolutional layer detect only the paddle. Conv2 filters begin to detect ball movement in particular directions and some jointly track the ball and the paddle. Nearly all Conv3 filters track ball and paddle interactions including deflections, ball velocity, and direction of travel. Despite seeing a single frame at a time, individual LSTM units also detect high level events, respectively: the agent missing the ball, ball reflections off of paddles, and ball reflections off the walls. Each image superimposes the last 10-frames seen by the agent, giving more luminance to the more recent frames.}
\label{fig:convMaxAct}
\end{figure}

Perhaps the most important opportunity presented by a history of game
screens is the ability to convolutionally detect object
velocity. Figure \ref{fig:convMaxAct} visualizes the game screens
maximizing the activations of different convolutional filters and
confirms that the 10-frame DQN's filters do detect object velocity,
though perhaps less reliably than normal unobscured
Pong.\footnote{\cite{guo14} also confirms that convolutional filters
learn to respond to patterns of movement seen in game objects.}

Remarkably, DRQN performs well at this task even when given only one
input frame per timestep. With a single frame it is impossible for
DRQN's convolutional layers to detect any type of velocity. Instead,
the higher-level recurrent layer must compensate for both the
flickering game screen and the lack of convolutional velocity
detection. Figure \ref{fig:lstm_units} confirms that individual units
in the LSTM layer are capable of integrating noisy single-frame
information through time to detect high-level Pong events such as the
player missing the ball, the ball reflecting on a paddle, or the ball
reflecting off the wall.

DRQN is trained using backpropagation through time for the last ten
timesteps. Thus both the non-recurrent 10-frame DQN and the recurrent
1-frame DRQN have access to the same history of game
screens.\footnote{However, \cite{karpathy15} show that LSTMs can learn
functions at training time over a limited set of timesteps and then
generalize them at test time to longer sequences.} Thus, when dealing
with partial observability, a choice exists between using a
non-recurrent deep network with a long history of observations or
using a recurrent network trained with a single observation at each
timestep. The results in this section show that recurrent networks can
integrate information through time and serve as a viable alternative
to stacking frames in the input layer of a convoluational network.

\section{Evaluation on Standard Atari Games}
We selected the following nine Atari games for
evaluation: \textit{Asteroids} and \textit{Double Dunk} feature
naturally-flickering sprites making them good potential candidates for
recurrent learning. \textit{Beam Rider}, \textit{Centipede},
and \textit{Chopper Command} are shooters. \textit{Frostbite} is a
platformer similar to Frogger. \textit{Ice Hockey} and \textit{Double
Dunk} are sports games that require positioning players, passing and
shooting the puck/ball, and require the player to be capable of both
offense and defense. \textit{Bowling} requires actions to be taken at
a specific time in order to guide the ball. \textit{Ms Pacman}
features flickering ghosts and power pills.

\begin{figure}[t!]
\begin{subfigure}{\linewidth}
  \includegraphics[width=\linewidth]{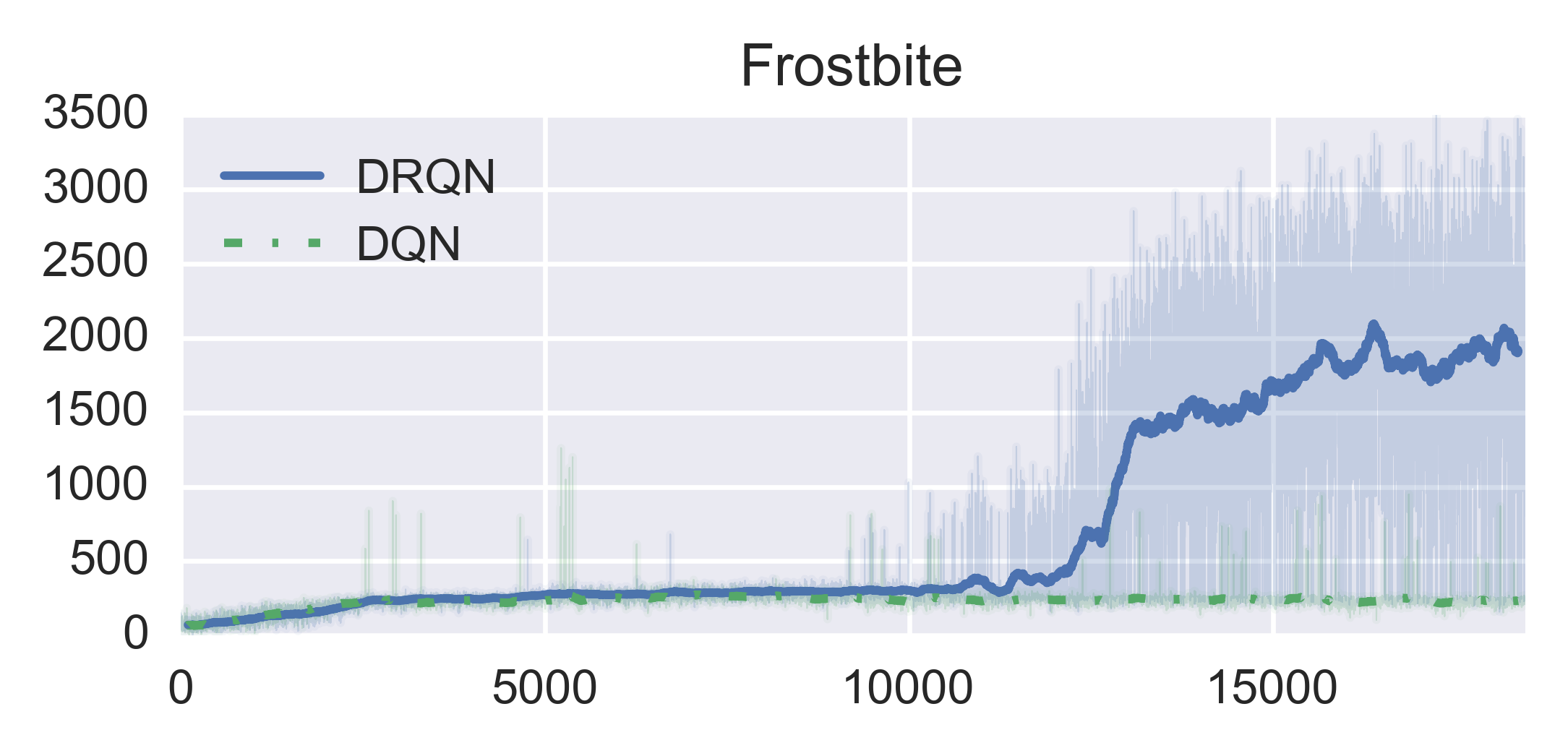}
\end{subfigure}
\begin{subfigure}{\linewidth}
  \includegraphics[width=\linewidth]{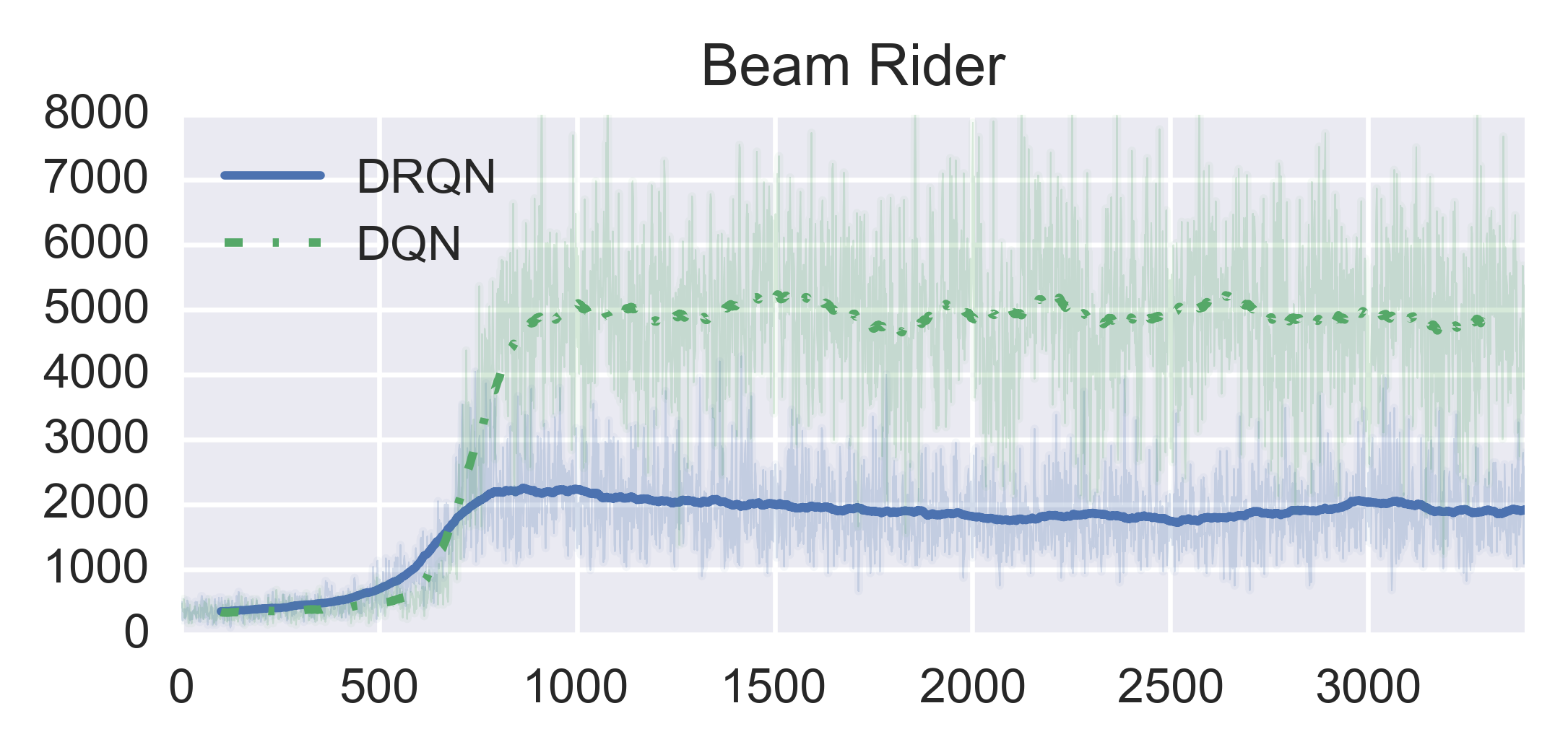}
\end{subfigure}
\caption{Frostbite and Beam Rider represent the best and worst games for DRQN. Frostbite performance jumps as the agent learns to reliably complete the first level.}
\label{fig:frostbeamrider}
\end{figure}

Given the last four frames of input, all of these games are MDPs
rather than POMDPs. Thus there is no reason to expect DRQN to
outperform DQN. Indeed, results in Table \ref{tab:baseline} indicate
that on average, DRQN does roughly as well DQN. Specifically, our
re-implementation of DQN performs similarly to the original,
outperforming the original on five out of the nine games, but
achieving less than half the original score on Centipede and Chopper
Command. DRQN performs outperforms our DQN on the games of Frostbite
and Double Dunk, but does significantly worse on the game of Beam
Rider (Figure \ref{fig:frostbeamrider}). The game of Frostbite
(Figure \ref{fig:frostbite}) requires the player to jump across all
four rows of moving icebergs and return to the top of the
screen. After traversing the icebergs several times, enough ice has
been collected to build an igloo at the top right of the
screen. Subsequently the player can enter the igloo to advance to the
next level. As shown in Figure \ref{fig:frostbeamrider}, after 12,000
episodes DRQN discovers a policy that allows it to reliably advance
past the first level of Frostbite. For experimental details, see
Appendix B.

\begin{table}[htp]
\small
\centering
\begin{tabular}{l l l | l}
  & DRQN $\pm std$ & \multicolumn{2}{c}{DQN $\pm std$} \\
  \hline
  Game &  & Ours & Mnih et al. \\
  \hline
  Asteroids & 1020 $(\pm312)$ & 1070 $(\pm345)$ & 1629 $(\pm542)$ \\
  Beam Rider & 3269 $(\pm1167)$ & \textbf{6923} $(\pm1027)$ & 6846 $(\pm1619)$\\
  Bowling & 62 $(\pm5.9)$ & 72 $(\pm11)$ & 42 $(\pm88)$\\
  Centipede & 3534 $(\pm1601)$ & 3653 $(\pm1903)$ & 8309 $(\pm5237)$\\
  Chopper Cmd & 2070 $(\pm875)$ & 1460 $(\pm976)$ & 6687 $(\pm2916)$\\
  Double Dunk & \textbf{-2} $(\pm7.8)$ & -10 $(\pm3.5)$ & -18.1 $(\pm2.6)$\\
  Frostbite & \textbf{2875} $(\pm535)$ & 519 $(\pm363)$ & 328.3 $(\pm250.5)$\\
  Ice Hockey & -4.4 $(\pm1.6)$ & -3.5 $(\pm3.5)$ & -1.6 $(\pm2.5)$\\
  Ms. Pacman & 2048 $(\pm653)$ & 2363 $(\pm735)$ & 2311 $(\pm525)$\\
\end{tabular}
\caption{On standard Atari games, DRQN performance parallels DQN, excelling in the games of Frostbite and Double Dunk, but struggling on Beam Rider. Bolded font indicates statistical significance between DRQN and our DQN.\footnotemark[5]}
\label{tab:baseline}
\end{table}

\footnotetext[5]{Statistical significance of scores determined by independent t-tests using Benjamini-Hochberg procedure and significance level $P=.05$.}

\section{MDP to POMDP Generalization}
\begin{figure}[htp]
  \centering
  \includegraphics[width=.5\textwidth]{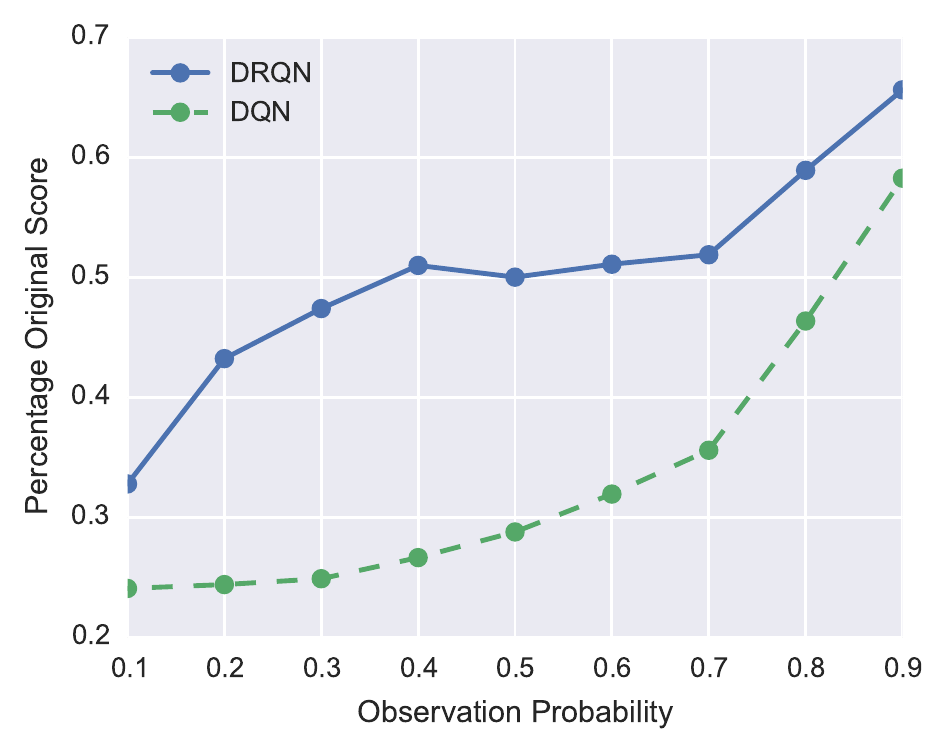}
\caption{When trained on normal games (MDPs) and then evaluated on flickering games (POMDPs), DRQN's performance degrades more gracefully than DQN's. Each data point shows the average percentage of the original game score over all 9 games in Table \ref{tab:baseline}.}
\label{fig:reverse_gen}
\end{figure}

Can a recurrent network be trained on a standard MDP and then
generalize to a POMDP at evaluation time? To address this question, we
evaluate the highest-scoring policies of DRQN and DQN over the
flickering equivalents of all 9 games in
Table \ref{tab:baseline}. Figure \ref{fig:reverse_gen} shows that
while both algorithms incur significant performance decreases on
account of the missing information, DRQN captures more of its previous
performance than DQN across all levels of flickering. We conclude that
recurrent controllers have a certain degree of robustness against
missing information, even trained with full state information.

\section{Related Work}
Previously, LSTM networks have been demonstrated to solve POMDPs when
trained using policy gradient methods \cite{wierstra07}. In contrast
to policy gradient, our work uses temporal-difference updates to
bootstrap an action-value function. Additionally, by jointly training
convolutional and LSTM layers we are able to learn directly from
pixels and do not require hand-engineered features.

LSTM has been used as an advantage-function approximator and shown to
solve a partially observable corridor and cartpole tasks better better
than comparable (non-LSTM) RNNs \cite{bakker01}. While similar in
principle, the corridor and cartpole tasks feature tiny states spaces
with just a few features.

In parallel to our work, \cite{narasimhan15} independently combined
LSTM with Deep Reinforcement Learning to demonstrate that recurrency
helps to better play text-based fantasy games. The approach is similar
but the domains differ: despite the apparent complexity of the
fantasy-generated text, the underlying MDPs feature relatively
low-dimensional manifolds of underlying state space. The more complex
of the two games features only 56 underlying states. Atari games, in
contrast, feature a much richer state space with typical games having
millions of different states. However, the action space of the text
games is much larger with a branching factor of 222 versus Atari's 18.

\section{Discussion and Conclusion}
Real-world tasks often feature incomplete and noisy state information,
resulting from partial observability. We modify DQN to handle the
noisy observations characteristic of POMDPs by combining a Long Short
Term Memory with a Deep Q-Network. The resulting \textit{Deep
Recurrent Q-Network} (DRQN), despite seeing only a single frame at
each step, is still capable integrating information across frames to
detect relevant information such as velocity of on-screen objects.
Additionally, on the game of Pong, DRQN is better equipped than a
standard Deep Q-Network to handle the type of partial observability
induced by flickering game screens.

Furthermore, when trained with partial observations, DRQN can
generalize its policies to the case of complete observations. On the
Flickering Pong domain, performance scales with the observability of
the domain, reaching near-perfect levels when every game screen is
observed. This result indicates that the recurrent network learns
policies that are both robust enough to handle to missing game screens
and scalable enough to improve performance as observability
increases. Generalization also occurs in the opposite direction: when
trained on standard Atari games and evaluated against flickering
games, DRQN's performance generalizes better than DQN's at all levels
of partial information.

Our experiments suggest that Pong represents an outlier among the
examined games. Across a set of ten Flickering MDPs we observe no
systematic improvement when employing recurrency. Similarly, across
non-flickering Atari games, there are few significant differences
between the recurrent and non-recurrent player. This observation leads
us to conclude that while recurrency is a viable method for handling
state observations, it confers no systematic benefit compared to
stacking the observations in the input layer of a convolutional
network. One avenue for future is identifying the relevant
characteristics of the Pong and Frostbite that lead to better
performance by recurrent networks.

\section{Acknowledgments}
This work has taken place in the Learning Agents Research Group (LARG)
at the Artificial Intelligence Laboratory, The University of Texas at
Austin.  LARG research is supported in part by grants from the
National Science Foundation (CNS-1330072, CNS-1305287), ONR
(21C184-01), AFRL (FA8750-14-1-0070), AFOSR (FA9550-14-1-0087), and
Yujin Robot. Additional support from the Texas Advanced Computing
Center, and Nvidia Corporation.

\bibliography{drqn}

\begin{thebibliography}{}

\bibitem[\protect\citeauthoryear{Bakker}{2001}]{bakker01}
Bakker, B.
\newblock 2001.
\newblock Reinforcement learning with long short-term memory.
\newblock In {\em NIPS},  1475--1482.
\newblock MIT Press.

\bibitem[\protect\citeauthoryear{{Bellemare} \bgroup et al\mbox.\egroup
  }{2013}]{bellemare13}
{Bellemare}, M.~G.; {Naddaf}, Y.; {Veness}, J.; and {Bowling}, M.
\newblock 2013.
\newblock The arcade learning environment: An evaluation platform for general
  agents.
\newblock {\em Journal of Artificial Intelligence Research} 47:253--279.

\bibitem[\protect\citeauthoryear{Cun \bgroup et al\mbox.\egroup
  }{1998}]{lecun98}
Cun, Y. L.~L.; Bottou, L.; Bengio, Y.; and Haffner, P.
\newblock 1998.
\newblock Gradient-based learning applied to document recognition.
\newblock {\em Proceedings of IEEE} 86(11):2278--2324.

\bibitem[\protect\citeauthoryear{Guo \bgroup et al\mbox.\egroup }{2014}]{guo14}
Guo, X.; Singh, S.; Lee, H.; Lewis, R.~L.; and Wang, X.
\newblock 2014.
\newblock Deep learning for real-time atari game play using offline monte-carlo
  tree search planning.
\newblock In Ghahramani, Z.; Welling, M.; Cortes, C.; Lawrence, N.; and
  Weinberger, K., eds., {\em Advances in Neural Information Processing Systems
  27}. Curran Associates, Inc.
\newblock  3338--3346.

\bibitem[\protect\citeauthoryear{Hochreiter and
  Schmidhuber}{1997}]{hochreiter97}
Hochreiter, S., and Schmidhuber, J.
\newblock 1997.
\newblock Long short-term memory.
\newblock {\em Neural Comput.} 9(8):1735--1780.

\bibitem[\protect\citeauthoryear{Jia \bgroup et al\mbox.\egroup }{2014}]{jia14}
Jia, Y.; Shelhamer, E.; Donahue, J.; Karayev, S.; Long, J.; Girshick, R.;
  Guadarrama, S.; and Darrell, T.
\newblock 2014.
\newblock Caffe: Convolutional architecture for fast feature embedding.
\newblock {\em arXiv preprint arXiv:1408.5093}.

\bibitem[\protect\citeauthoryear{Karpathy, Johnson, and Li}{2015}]{karpathy15}
Karpathy, A.; Johnson, J.; and Li, F.-F.
\newblock 2015.
\newblock Visualizing and understanding recurrent networks.
\newblock {\em arXiv preprint}.

\bibitem[\protect\citeauthoryear{Mnih \bgroup et al\mbox.\egroup
  }{2015}]{mnih15}
Mnih, V.; Kavukcuoglu, K.; Silver, D.; Rusu, A.~A.; Veness, J.; Bellemare,
  M.~G.; Graves, A.; Riedmiller, M.; Fidjeland, A.~K.; Ostrovski, G.; Petersen,
  S.; Beattie, C.; Sadik, A.; Antonoglou, I.; King, H.; Kumaran, D.; Wierstra,
  D.; Legg, S.; and Hassabis, D.
\newblock 2015.
\newblock {Human-level control through deep reinforcement learning}.
\newblock {\em Nature} 518(7540):529--533.

\bibitem[\protect\citeauthoryear{Narasimhan, Kulkarni, and
  Barzilay}{2015}]{narasimhan15}
Narasimhan, K.; Kulkarni, T.; and Barzilay, R.
\newblock 2015.
\newblock Language understanding for text-based games using deep reinforcement
  learning.
\newblock {\em CoRR} abs/1506.08941.

\bibitem[\protect\citeauthoryear{Sutton and Barto}{1998}]{sutton98}
Sutton, R.~S., and Barto, A.~G.
\newblock 1998.
\newblock {\em Reinforcement Learning: An Introduction}.
\newblock {MIT} Press.

\bibitem[\protect\citeauthoryear{Tieleman and Hinton}{2012}]{tieleman12}
Tieleman, T., and Hinton, G.
\newblock 2012.
\newblock {Lecture 6.5---RmsProp: Divide the gradient by a running average of
  its recent magnitude}.
\newblock COURSERA: Neural Networks for Machine Learning.

\bibitem[\protect\citeauthoryear{Tsitsiklis and Roy}{1997}]{tsitsiklis97}
Tsitsiklis, J.~N., and Roy, B.~V.
\newblock 1997.
\newblock An analysis of temporal-difference learning with function
  approximation.
\newblock {\em IEEE Transactions on Automatic Control} 42(5):674--690.

\bibitem[\protect\citeauthoryear{Watkins and Dayan}{1992}]{watkins92}
Watkins, C. J. C.~H., and Dayan, P.
\newblock 1992.
\newblock Q-learning.
\newblock {\em Machine Learning} 8(3-4):279--292.

\bibitem[\protect\citeauthoryear{Wierstra \bgroup et al\mbox.\egroup
  }{2007}]{wierstra07}
Wierstra, D.; Foerster, A.; Peters, J.; and Schmidthuber, J.
\newblock 2007.
\newblock Solving deep memory {POMDP}s with recurrent policy gradients.

\bibitem[\protect\citeauthoryear{Zeiler}{2012}]{zeiler12}
Zeiler, M.~D.
\newblock 2012.
\newblock {ADADELTA}: An adaptive learning rate method.
\newblock {\em CoRR} abs/1212.5701.

\end{thebibliography}
\bibliographystyle{aaai}

\newpage
\appendix

\section{Appendix A: Alternative Architectures}
Several alternative architectures were evaluated on the game of Beam
Rider. We explored the possibility of either replacing the first
non-convolutional fully connected layer with an LSTM layer (LSTM
replaces IP1) or adding the LSTM layer between the first and second
fully connected layers (LSTM over IP1). Results strongly indicated
LSTM should replace IP1. We hypothesize this allows LSTM direct access
to the convolutional features. Additionally, adding a Rectifier layer
after the LSTM layer consistently reduced performance.

\begin{table}[htp]
\begin{tabular}{l r r }
  \hline
  Description & Percent Improvement \\
  \hline
  LSTM replaces IP1 & 709\% \\
  ReLU-LSTM replaces IP1 & 533\% \\
  LSTM over IP1 & 418\% \\
  ReLU-LSTM over IP1 & 0\% \\
\end{tabular}
\end{table}

Another possible architecture combines frame stacking from DQN with
the recurrency of LSTM. This architecture accepts a stack of the four
latest frames at every timestep. The LSTM portion of the architecture
remains the same and is unrolled over the last 10 timesteps. In
theory, this modification should allow velocity detection to happen in
the convolutional layers of the network, leaving the LSTM free to
perform higher-order processing. This architecture has the largest
number of parameters and requires the most training
time. Unfortunately, results show that the additional parameters do
not lead to increased performance on the set of games examined. It is
possible that the network has too many parameters and is prone to
overfitting the training experiences it has seen.

\section{Appendix B: Computational Efficiency}
Computational efficiency of RNNs is an important concern. We conducted
experiments by performing 1000 backwards and forwards passes and
reporting the average time in milliseconds required for each
pass. Experiments used a single Nvidia GTX Titan Black using CuDNN and
a fully optimized version of Caffe. Results indicate that computation
scales sub-linearly in both the number of frames stacked in the input
layer and the number of iterations unrolled. Even so, models trained
on a large number of stacked frames and unrolled for many iterations
are often computationally intractable. For example a model unrolled
for 30 iterations with 10 stacked frames would require over 56 days to
reach 10 million iterations.

\begin{table}[htp]
\begin{tabular}{l r r r| r r r}
  & \multicolumn{3}{c|}{Backwards (ms)} & \multicolumn{3}{c}{Forwards (ms)} \\
  \hline
  Frames & 1 & 4 & 10 & 1 & 4 & 10 \\
  \hline
  Baseline & 8.82 & 13.6 & 26.7 & 2.0 & 4.0 & 9.0 \\
  Unroll 1 & 18.2 & 22.3 & 33.7 & 2.4 & 4.4 & 9.4 \\
  Unroll 10 & 77.3 & 111.3 & 180.5 & 2.5 & 4.4 & 8.3 \\
  Unroll 30 & 204.5 & 263.4 & 491.1 & 2.5 & 3.8 & 9.4 \\
\end{tabular}
\caption{Average milliseconds per backwards/forwards pass. Frames
  refers to the number of channels in the input image. Baseline is a
  non recurrent network (e.g. DQN). Unroll refers to an LSTM network
  backpropagated through time $1/10/30$ steps.}
\end{table}

\section{Appendix C: Experimental Details}
Policies were evaluated every 50,000 iterations by playing 10 episodes
and averaging the resulting scores. Networks were trained for 10
million iterations and used a replay memory of size
400,000. Additionally, all networks used ADADELTA \cite{zeiler12}
optimizer with a learning rate of 0.1 and momentum of 0.95. LSTM's
gradients were clipped to a value of ten to ensure learning
stability. All other settings were identical to those given
in \cite{mnih15}.

All networks were trained using the Arcade Learning Environment
ALE \cite{bellemare13}. The following ALE options were used: color
averaging, minimal action set, and death detection.

DRQN is implemented in Caffe \cite{jia14}. The source is available
at [removed for blind review].

\section{Appendix C: Flickering Results}
\begin{table}[htp]
\small
\centering
\begin{tabular}{l l l}
  \hline
  Flickering & DRQN $\pm std$ & DQN $\pm std$ \\
  \hline
  Asteroids & 1032 $(\pm410)$ & 1010 $(\pm535)$ \\
  Beam Rider & 618 $(\pm115)$ & \textbf{1685.6} $(\pm875)$ \\
  Bowling & 65.5 $(\pm13)$ & 57.3 $(\pm8)$ \\
  Centipede & 4319.2 $(\pm4378)$ & 5268.1 $(\pm2052)$ \\
  Chopper Cmd & 1330 $(\pm294)$ & 1450 $(\pm787.8)$ \\
  Double Dunk & -14 $(\pm2.5)$ & -16.2 $(\pm2.6)$ \\
  Frostbite & 414 $(\pm494)$ & 436 $(\pm462.5)$ \\
  Ice Hockey & -5.4 $(\pm2.7)$ & -4.2 $(\pm1.5)$ \\
  Ms. Pacman & 1739 $(\pm942)$ & 1824 $(\pm490)$ \\
  Pong & \textbf{12.1} $(\pm2.2)$ & -9.9 $(\pm3.3)$ \\
\end{tabular}
\caption{Each screen is obscured with probability 0.5, resulting in a partially-observable, flickering equivalent of the standard game. Bold font indicates statistical significance.\footnotemark[5]}
\label{tab:flickering_results}
\end{table}

\end{document}